\documentclass[runningheads]{llncs}
\usepackage[T1]{fontenc}
\usepackage{graphicx}
\usepackage{helpers/style}
\begin{document}
%
%
%
%
\maketitle              
\begin{abstract}

Time-series anomaly detection (TSAD) is critical in domains such as industrial monitoring, healthcare, and cybersecurity, but it remains challenging due to rare and heterogeneous anomalies and the scarcity of labelled data.
This scarcity makes unsupervised approaches predominant, yet existing methods often rely on reconstruction or forecasting, which struggle with complex data, or on embedding-based approaches that require domain-specific anomaly synthesis and fixed distance metrics.
We propose \ours{}, a framework that generates pseudo-anomalies directly in the latent space, avoiding handcrafted anomaly injections and the need for domain expertise.
A latent-space decoder produces tailored pseudo-anomalies to train a Transformer-based anomaly classifier, while a pre-trained LLM enriches the temporal and contextual representations of this space.
Experiments on three benchmark datasets show that \ours{} achieves state-of-the-art performance and sets a new standard for LLM-based TSAD.

\end{abstract}
\begingroup
\renewcommand\thefootnote{}
\footnotetext{\href{https://gitlab.com/uniluxembourg/snt/cvi2/open/space/aster-tab}{https://gitlab.com/uniluxembourg/snt/cvi2/open/space/aster-tab}}
\endgroup

\ifSubfilesClassLoaded\maketitle

\section{Introduction}

Time-series anomaly detection (TSAD) focuses on identifying patterns or behaviours in temporal data that deviate from expected dynamics.
It plays a critical role in a wide range of real-world applications, including industrial fault diagnosis~\cite{mathur2016swat}, financial fraud detection~\cite{dal2015calibrating}, healthcare monitoring~\cite{greenwald1990improved}, and cybersecurity~\cite{filonov2016multivariate}.
Unlike anomaly detection on static supports, such as images or 3D data~\cite{Hermary2025RemovingGB,ruff2018deep,Gu2024RethinkingRD}, TSAD must account for temporal dependencies between observations and evolving data distributions, making the task considerably more complex.
Specifically, detecting anomalies in time-series is inherently challenging due to the rarity of anomalous events, their heterogeneous temporal patterns, and their dependence on contextual information, requiring models to account for both temporal dynamics and interactions across multiple features.

These challenges are compounded by the scarcity of labelled anomalies in practical scenarios, motivating the development of unsupervised approaches\,\allowbreak\cite{tuli2022tranad,ky2024cats} that can effectively model normal behaviour and detect deviations without reliance on annotated data.

Among unsupervised methods, we find embedding-based strategies~\cite{ky2024cats,kim2023contrastive}.
They aim to learn a transformation that maps normal patterns to a dense region in the representation space, effectively separating them from abnormal patterns.
In this paradigm, we commonly find approaches that aim to maximise the distance between the embeddings of normal data and synthetically generated anomalies--or \textit{pseudo-anomalies}.
For these methods, the accurate and realistic generation of synthetic anomalies is crucial for training a reliable anomaly detection model.
To this end, most rely on domain knowledge to tediously curate lists of possible anomaly types.
These selective lists are then used as a basis for defining transformations, such as adding bias, noise, or masking.
These corruptive transformations can then be applied to normal data to generate pseudo-anomalies.
However, considering the nature of the multi-variate TSAD problem, this procedural, data augmentation-dependent anomaly generation is confined to the curated lists and restricts the performance and generalisation of the models.

An alternative direction could involve making the transformations adapt automatically to the data by \textit{learning} them; and to further improve generalisation, applying them within a less domain-specific space, \ie, a latent representation space.
This, however, requires the ability to construct expressive latent representations that capture temporal and cross-feature dynamics.
Unfortunately, robust foundational models for time-series remain scarce, owing to data heterogeneity and limited availability~\cite{chang2023llm4ts}.
Recently, Large Language Models (LLMs) have been adapted to solve various time-series problems, such as forecasting~\cite{jin2024timellm}, classification~\cite{zhou2023one}, and anomaly detection~\cite{tao2025madllm}.
These pre-trained models have proven to transfer helpful cross-domain and cross-modality knowledge thanks to their strong representation capabilities~\cite{Lu2022FrozenPT,Sui2024TableML}.
Additionally, given the rapid advancements and expanding capabilities of LLMs, formulating a TSAD method adapted to their architecture represents a valuable research opportunity.
In this work, we introduce \ours, \textit{Anomaly Synthesis through Transformer-based Embedding Representation}, a novel framework for unsupervised TSAD.
\ours{} generates pseudo-anomalies directly in latent space via learnt transformations, avoiding handcrafted, domain-specific augmentations in the raw data space.
A pre-trained LLM is first used to encode time-series windows into rich, contextualised embeddings.
These embeddings are then processed by a VAE-based module, the \textit{perturbator}, which synthesises diverse and non-trivial pseudo-anomalies.
Unlike metric-based approaches that depend on fixed distance measures, \ours{} trains a Transformer-based classifier to learn a flexible decision boundary.
This design enables \ours{} to capture complex abnormal behaviours without relying on predefined anomaly types.

Our results show that \ours{} achieves a new state-of-the-art performance on multiple datasets and improves upon the previous best LLM-based methods. We summarise our contributions as follows:%
\begin{itemize}
    \item We propose an unsupervised TSAD framework that generates pseudo\hyp anomalies in latent space, eliminating domain-specific augmentations and improving generalisation.
    \item We introduce ASTER, a novel architecture featuring a VAE-based \textit{perturbator} that models normal data and learns to generate pseudo-anomalies.
    It is paired with a Transformer-based classifier that learns a discriminative decision boundary.
    \item We demonstrate how pre-trained LLMs can be effectively leveraged as contextual feature extractors for TSAD.
    \item We validate our method using the rigorous and unified TAB benchmark~\cite{qiu2025tab}.
    This addresses inconsistencies in prior evaluations by enabling fair and reproducible comparisons across diverse TSAD methods.
\end{itemize}

The remainder of the paper is organised as follows: we provide an extensive review of related works in \Cref{sec:related}.
Next, we formally introduce the problem and the proposed method in \Cref{sec:formul} and \Cref{sec:method}, respectively. 
Our results and ablation studies are presented in \Cref{sec:exp}.
We then discuss certain limitations and directions for future work in \Cref{sec:discussion}, and conclude in \Cref{sec:conc}.

\ifSubfilesClassLoaded{%
\bibliographystyle{ACM-Reference-Format}%
\bibliography{bib,general}%
}

\end{document}

\ifSubfilesClassLoaded\maketitle

\section{Related Work}
\label{sec:related}

In this section, we first provide a summary of TSAD methods, then focus on methods that actively refine embeddings of time-series data to reflect temporal patterns and elaborate on the adoption of LLMs for various time-series problems.
Finally, we explain the efforts to provide a comprehensive benchmark for a fair evaluation and comparison of TSAD methods. 

\subsection{Time-Series Anomaly Detection}

Early TSAD research relied on classical statistical techniques~\cite{Moayedi2008ArimaMF} and traditional machine learning methods, including autoregressive models, hypothesis testing, and distance-, density-, or isolation-based approaches~\cite{chaovalitwongse2007time,breunig2000lof,liu2008isolation}.
However, these methods often struggle with complex temporal dependencies and non-linear patterns in high-dimensional data.

Deep learning has enabled models that learn expressive representations from raw inputs~\cite{ruff2018deep}.
LSTM-based models~\cite{hundman2018detecting}, VAEs~\cite{yao2023regularizing}, and diffusion-based methods~\cite{chen2023imdiffusion} improve reconstruction of normal patterns, while Generative Adversarial Networks (GANs)~\cite{zenati2018adversarially,zhou2019beatgan} introduce adversarial anomaly synthesis, and one-class classifiers like Deep SVDD~\cite{ruff2018deep} enhance robustness to noisy pseudo-labels.

Unsupervised paradigms dominate due to a scarcity of labelled anomalies.
Reconstruction-, forecasting-, and imputation-based~\cite{audibert2020usad,tan2024language,xiao2023imputation} methods detect anomalies through deviations from expected patterns.
Self-supervised techniques, such as contrastive learning, masked modelling, or transformation prediction~\cite{ts2vec}, further improve latent representations.
Transformer-based models~\cite{tuli2022tranad,liu2024timer,Wen2022TransformersIT} capture long-range dependencies, improving performance on multivariate time-series.

\subsection{Embedding-based TSAD}
Shaping appropriate data embeddings suggests extracting informative features that capture the underlying data structures without task-specific supervision \cite{bengio2013representation,lecun2015deep}.
This is, in fact, closely related to the field of representation learning~\cite{ts2vec}.

Contrastive learning~\cite{oord2018representation,chen2020simple} has emerged as a powerful paradigm for unsupervised TSAD~\cite{darban2025carla}.
By maximising alignment between augmented views of a sample (or similar ones) while distancing others, these models learn discriminative embeddings without labels.
Contrastive TSAD methods, such as CTAD~\cite{kim2023contrastive} and CATS~\cite{ky2024cats}, rely on specific augmentations (\eg, masking, additive bias) to generate synthetic anomalies as negative samples, maximising their distance from normal representations using the contrastive loss.
However, these strategies often require dataset-specific hyper-parameter tuning, limiting generalisation.
CARLA~\cite{darban2025carla} partially addresses this by confining the augmentation-based generation to a pre-training stage.
Despite their success, these methods remain limited by manual, domain-intensive augmentation design in raw input space.

Latent space exploration, on the other hand, has traditionally focused on modelling normal data distributions.
A prominent example is DAGMM~\cite{Zong2018DeepAG,bhatnagar2021merlion}, which utilises Gaussian Mixture Models (GMMs) in the latent space to identify low-probability deviations as anomalies.
Similarly, TimeVQVAE-AD~\cite{lee2024explainable} employs a two-stage time-frequency auto-encoder to detect anomalies based on estimated densities and reconstruction quality.
An increasing interest in latent space manipulation is also noticeable; for instance, L-GTA \cite{roque2025gta} applies augmentations directly in the latent space to enhance reconstruction robustness.

Departing from these approaches, our method does not seek to improve reconstruction performance, nor do we rely on passive density modelling of normal data.
Instead, we leverage the latent space to model a specific distribution for pseudo-anomalies.
By sampling directly from this learnt latent distribution, we circumvent the need for heuristic-driven augmentations in the raw input space, which are often suboptimal and domain-dependent.
This strategy, concurrently explored in GenIAS~\cite{Darban2025GenIASGF}, facilitates a more versatile, dataset-agnostic representation that removes the requirement for extensive domain knowledge.
Consequently, the latent space and pseudo-anomaly generation process are optimised for the downstream task of anomaly detection.

\begin{figure*}[t]
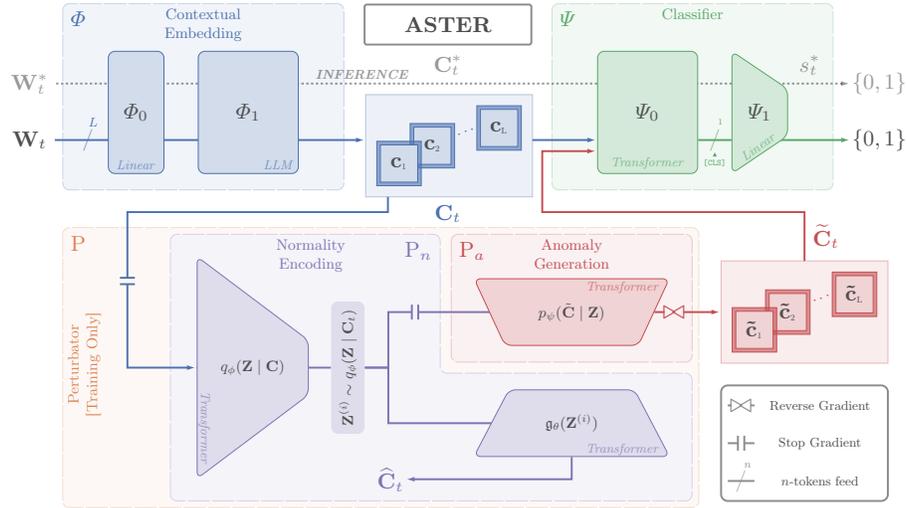

    \centering
    \resizebox{\linewidth}{!}{\subfile{../figures/tikz/overall-lined}}%
   
    \caption{Overview of the proposed method. It consists of three components: Contextual Embedding ($\Phi$) extractor, Perturbator ($\textrm{P}$), and Classifier ($\Psi$). The Contextual Embedding maps the input time series to a high-dimensional space using a pre-trained LLM. The Perturbator learns a latent representation via reconstruction ($\textrm{P}_n$) and generates pseudo-anomalies ($\textrm{P}_a$). The Classifier distinguishes between normal and pseudo-anomalous data.}
    \label{fig:overall}
\end{figure*}
\subsection{LLMs \& Time-Series}

Recently, Large Language Models (LLMs)~\cite{radford2019language,devlin2019bert}, which have excelled at capturing sequential relations in text data, have been adopted in the time-series domain to solve various tasks, such as forecasting~\cite{gruver2023large,chang2023llm4ts,tan2024language,sun2024test,jin2024timellm}, reasoning~\cite{chatts}, and anomaly detection~\cite{zhou2023one,liu2024large2,alnegheimish2024sigllm,tao2025madllm,liu2025large,zhou2025can}.

GPT4TS~\cite{zhou2023one} is an early work that adapts a pre-trained GPT-2~\cite{radford2019language} model for time-series tasks, including anomaly detection, using lightweight fine-tuning.
Following this direction, UniTime~\cite{liu2024unitime} introduces masking-based input processing and incorporates domain knowledge via textual descriptions. AnomalyLLM~\cite{liu2024large2} adopts a knowledge distillation framework, where a Transformer-based student model learns from a GPT-2-initialised teacher, leveraging reconstruction errors for anomaly detection.

Another line of work explores the zero-shot capabilities of LLMs for time-series anomaly detection.
SIGLLM~\cite{alnegheimish2024sigllm} formulates anomaly detection as either direct prompting or forecasting-based discrepancy estimation.
LLMAD~\cite{liu2025large} employs retrieval-augmented in-context learning to mimic expert reasoning, while AnomLLM~\cite{zhou2025can} systematically analyses common assumptions in LLM-based TSAD, showing that longer input windows and Chain-of-Thought prompting do not consistently improve performance.

Motivated by these findings, we leverage a pre-trained LLM as a contextual feature extractor with minimal fine-tuning to bridge domain gaps, without introducing additional textual prompts or expert-defined context.
\subsection{Benchmarking TSAD Models}
Fair benchmarking of TSAD methods is crucial for accurately assessing progress.
Methods achieving state-of-the-art performance under specific protocols or metrics, \eg, point-adjusted $\textrm{F}_1$-score, often degrade under more realistic criteria such as the standard $\textrm{F}_1$-score.
Several benchmarking efforts have been proposed for univariate and multivariate TSAD~\cite{10.14778/3632093.3632110,Boniol2022TheseusNT}, but they largely exclude emerging paradigms, including LLM-based approaches and pre-training-based foundation models.
Recently, TAB~\cite{qiu2025tab} introduced a unified benchmark that standardises preprocessing, training, and evaluation, enabling fair and reproducible comparisons.
We therefore adopt TAB and its baselines for all evaluations.

\ifSubfilesClassLoaded{%
\bibliographystyle{ACM-Reference-Format}%
\bibliography{bib,general}%
}

\end{document}

\ifSubfilesClassLoaded\maketitle

\section{Problem Formulation}
\label{sec:formul}

Let $\fset{X}^0:=\{\fvm{x}_t\in\mathds{R}^D\}_{t=1}^T$ be a multivariate time-series of length $T$, with corresponding labels $\fset{Y}^0:=\{y_t\in\{0,1\}\}_{t=1}^T$, indicating normal and anomalous observations.
The goal of TSAD is to learn a function $\ffunc{f}$ that assigns an anomaly score $s_t=\ffunc{f}(\cdot)$ to each observation.
Since anomalies are context-dependent, $\ffunc{f}$ operates on a window of the past $L$ observations, \ie,
$s_t=\ffunc{f}([\fvm{x}_{t-L+1},\dots,\fvm{x}_t])$, where $\ffunc{f}:\mathds{R}^{L\times D}\to\mathds{R}$.
We thus process the time-series as a set of overlapping \textit{windows} $\fset{W}^0:=\{\fvm{W}_t\}_{t=L}^T$, with $\fvm{W}_t:=[\fvm{x}_{t-L+1},\dots,\fvm{x}_t]$.

At inference, an unseen time series $\fset{X}^*:=\{\fvm{x}_t\}_{t=1}^{T'}$, drawn from the same domain as $\fset{X}^0$, is used for evaluation.
Anomaly scores are computed sequentially for each window in $\fset{W}^*$, yielding $\fset{S}^*:=\{s^*_t\}_{t=L}^{T'}$.
A decision threshold $\tau\in\mathds{R}$ is then applied to obtain predicted labels $\widehat{\fset{Y}}^*:=\{\Hat{y}^*_t\}_{t=L}^{T'}$, where $\Hat{y}^*_t=\mathds{I}[s^*_t\ge\tau]\in\{0,1\}$ indicates an anomalous time-step.
Finally, comparing $\widehat{\fset{Y}}^*$ with the ground-truth labels $\fset{Y}^*$ provides the model’s performance on the unseen series.

The task is challenging for two reasons: anomaly detection is context\hyp dependent, and the setting is unsupervised ($y_t=0,\;\forall t$), making learning $\ffunc{f}$ non-trivial.
To prevent degenerate solutions, we generate a pseudo-anomalous window $\widetilde{\fvm{W}}_t$ for each input $\fvm{W}_t$, with the challenge being to create meaningful windows.

\ifSubfilesClassLoaded\clearpage

\end{document}

\ifSubfilesClassLoaded\maketitle

\section{Methodology}
\label{sec:method}

We present our architecture in \Cref{fig:overall}. The final objective $\ffunc{f}$ combines a contextual embedding $\Phi: \mathds{R}^{L\times D} \to \mathds{R}^{L\times M}$ (\Cref{subsec:embed}) and a classifier $\Psi: \mathds{R}^{L\times M} \to \mathds{R}$ (\Cref{subsec:class}), so that for a previously unseen window $\fvm{W}^*_t$ the anomaly score is $s^*_t = \Psi \circ \Phi(\fvm{W}^*_t)$.
To train $\Psi$ effectively, we introduce a pseudo-anomaly generator, the \textit{perturbator} $\text{P}: \mathds{R}^{L\times M} \to \mathds{R}^{L\times M}$, which produces plausible artificial anomalies (\Cref{subsec:pert}).

\subsection{Time-Series to Contextualised Embeddings}
\label{subsec:embed}

Anomalies in time-series are highly context-dependent, motivating the use of LLMs to capture rich sequential patterns.
This is the role of $\Phi$, which we divide into two parts: $\Phi_0: \mathds{R}^D \to \mathds{R}^M$, a time-step embedding, and $\Phi_1: \mathds{R}^{L\times M} \to \mathds{R}^{L\times M}$, which adds sequential context.

\textbf{\textit{Translation.}}
Observations are mapped from the raw time-series to the token space via a linear projection, which has been shown effective in prior works~\cite{zhou2023one,liu2025calf,Rukhovich2024CADRecodeRE}. This step only embeds individual time-steps, not cross-temporal knowledge:
\begin{equation}
\Phi_0: \mathds{R}^D\to \mathds{R}^M\qquad\qquad \Phi_0(\fvm{W}_t^{[l]}),\quad l \in t- L +1\dots t
\end{equation}

\textbf{\textit{LLM Knowledge.}}
The LLM incorporates the entire window with prior sequential knowledge, yielding a contextualised window $\fvm{C}_t \in \mathds{R}^{L \times M}$ when combined with the translation module (Equation~\ref{eq:contextwindow}).
$\Phi_0$ is trained from scratch, while $\Phi_1$ is only fine-tuned with LoRA~\cite{lora}, using gradients from the classification loss on normal windows $\nabla_\Phi = \frac{\partial\ffunc{L}_{CE}}{\partial\fvm{C}_t}$.
This keeps part of the network independent of pseudo-anomaly generation, reducing the risk of collapse or overfitting.

\begin{equation}
\label{eq:contextwindow}
\qquad \fvm{C}_t = \Phi(\fvm{W}_t) = \Phi_1 \circ \left( \left[\,\Phi_0(\fvm{W}_j^{[l]})\,\right]_{l=t- L +1}^t \right).
\end{equation}

\subsection{Classifier}
\label{subsec:class}

For the classifier $\Psi$, we follow a BERT-like strategy~\cite{devlin2019bert}. The contextualised tokens $\fvm{C}_t$ are first processed by $\Psi_0:\mathds{R}^{L\times M} \to \mathds{R}^{L\times M}$ (Transformer-based network) to gather contextual information into a \texttt{[CLS]} token. This token is selected using $\Pi_1(\fvm{X}) := [\fvm{X}]_{:,1}$ and projected to a final score by $\Psi_1: \mathds{R}^{M} \to \mathds{R}$:
\begin{equation}
\Psi := \Psi_1 \circ \Pi_1 \circ \Psi_0, \qquad s_t = \Psi(\fvm{C}_t).
\end{equation}
Applying a sigmoid, $\Bar{s}_t = \frac{1}{1 + e^{-s_t}}$, allows optimisation with the binary cross-entropy loss (Equation~\ref{eq:classloss}). Together with the contextual embedding $\Phi$, $\Psi$ completes the required components for the inference pipeline of \ours{}.

\tikz[remember picture, overlay]{\node[anchor=north west, xshift=.17\linewidth, rounded corners=1mm, yshift=.2cm, draw, minimum height=1cm, minimum width=7cm, line width=1.5pt, opacity=.8] {};}%
\setlength{\abovedisplayskip}{12pt} %
\setlength{\belowdisplayskip}{12pt} %
\begin{empheq}{align}
    \label{eq:classloss}
    \ffunc{L}_\textrm{CE} = -\left[y_{t}\mathrm{log}(\Bar{s}_{t}) + (1 - y_t)\mathrm{log}(1 - \Bar{s}_{t})\right].
\end{empheq}

\begin{figure}[t]
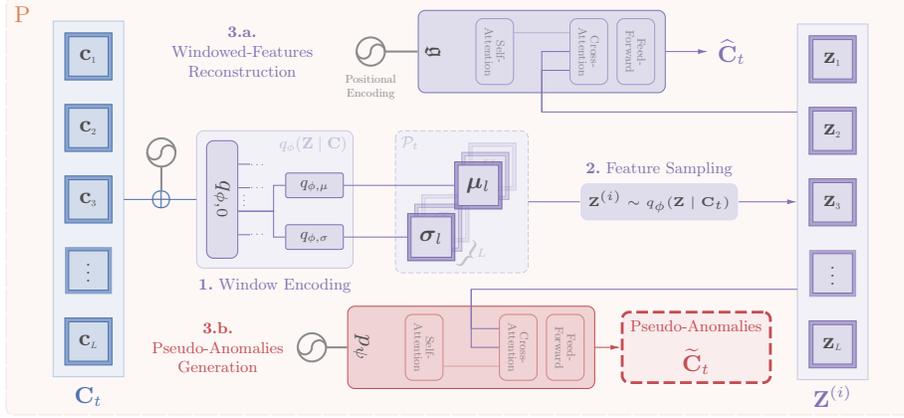

    \centering
    \resizebox{\linewidth}{!}{\subfile{../figures/tikz/perturbator}}%
    \caption{%
Detailed description of the Perturbator architecture.
It includes a complete Variational Auto-Encoder ({\color{color4}purple}), with an encoder $q_{\phi}$ that maps the input features $\fvm C_t$ to latent matrix $\fvm Z^{(i)}$, and a decoder $\ffunc g_\theta$.
The pseudo-anomalies are generated using the additional decoder $p_{\psi}$ ({\color{color3}red}), starting from the same latent matrix.
We also use positional encoding added to the inputs of each sub-network.
}
    \label{fig:pertubator}
\end{figure}

\subsection{Perturbator}
\label{subsec:pert}

As mentioned in \Cref{sec:method}, during training we use a perturbator $\textrm{P}$ to generate pseudo-anomalous samples and balance the data.
We establish three essential requirements for the perturbator:

\textit{1. Adaptive generation.} The process should automatically capture the structure of the data.
Current methods rely on fixed transformations (\eg, noise, shuffling, masking), which perform inconsistently across heterogeneous datasets.
A robust method must generalise without dataset-specific tuning;

\textit{2. Non-trivial anomalies.} While a simple alteration, such as adding random noise, suffices to simulate anomalies, these pseudo-anomalies are obvious when compared to the subtle, context-dependent nature of real anomalies.
Pseudo-anomalies should remain correlated with normal data while being hard to detect;

\textit{3. Diverse anomalies.} A reliable method must generalise across a wide range of anomalies to be effective.
A few realistic examples are insufficient to ensure robustness against arbitrary cases.

We address (1) and (2) by learning a distribution for anomalous windows, which enables automatic adaptation and estimation of anomaly difficulty.
To reduce bias from the original data, we generate pseudo-anomalies directly in the contextualised space.
We learn and model their distribution, $p(\Tilde{\fvm{C}})$, based on the estimated density of normal windows $p(\fvm{C})$.
To satisfy (3), we use a Variational Auto-Encoder (VAE)~\cite{kingma2013auto}, which provides a simple latent space and inherent diversity via their auxiliary variable $\bm \epsilon \sim \mathcal{N}(0,I)$.
Our approach models $p(\fvm{C})$ with an auto-encoding strategy (\Cref{subsec:modellingpc}) and uses its latent space to generate pseudo-anomalies (\Cref{subsec:anomalies}).
A detailed overview is presented in \Cref{fig:pertubator}.

\subsubsection{Modelling $p(\fvm{C})$.}
\label{subsec:modellingpc}

The core of VAE passes through a latent variable $\fvm Z$ and two distribution estimations $q_\phi(\fvm Z\mid\fvm C)$ and $p_\theta(\fvm C)$, parametrised by $\phi$ and $\theta$ respectively.
By defining $p_\theta(\fvm C)$ with its marginal over $\fvm Z$, $\int p_\theta(\fvm{C}, \fvm{Z}) \, d\fvm{Z}$, we can write the relation between $\log p_\theta(\fvm C)$ and the \emph{Evidence Lower Bound} (ELBO) as:
\begin{align}
\log p_\theta(\fvm{C}) 
&\geq 
\mathds{E}_{q_\phi(\fvm{Z} \mid \fvm{C})} \big[ \log p_\theta(\fvm{C} \mid \fvm{Z}) \big] - \ffunc{D}_{\mathrm{KL}} \big( q_\phi(\fvm{Z} \mid \fvm{C}) \, \| \, p(\fvm{Z}) \big) \\
&\geq \mathrm{ELBO}(\theta,\phi;\fvm{C}).
\end{align}
Maximising the ELBO with respect to the parameters $\theta$ and $\phi$ yields a tractable optimisation objective that approximates the true maximum likelihood estimation of the underlying data distribution $p(\fvm{C})$.
We thus obtain the optimisation objective for the modelling of $p(\fvm C)$:
\begin{align}
\ffunc{L}_\mathrm{ELBO} &:=  -\mathds{E}_{q_\phi(\fvm Z \mid \fvm C)}[\log p_\theta(\fvm C \mid \fvm Z)] + \ffunc{D}_{\mathrm{KL}}\big(q_\phi(\fvm Z \mid \fvm C) \,\|\, p(\fvm Z)\big).
\label{eq:elbo_gauss0}
\end{align}

A key modelling choice in VAEs is to consider $p_\theta(\fvm C\mid \fvm Z)=\mathcal N(\fvm C;\;\ffunc g_\theta(\fvm Z), \fvm I)$, where $\ffunc g_\theta : \mathds R^M \to \mathds R^M$ is the \textit{reconstruction} function.
This assumption allows to reduce the previous equation to:
\begin{align}
\ffunc{L}_\mathrm{ELBO} &\equiv \mathds{E}_{q_\phi(\fvm Z\mid\fvm C)}\!\big[\|\fvm C - \ffunc g_\theta(\fvm Z)\|^2\big] 
+ \ffunc{D}_{\mathrm{KL}}\big(q_\phi(\fvm Z\mid\fvm C)\,\|\,p(\fvm Z)\big).
\label{eq:elbo_gauss1}
\end{align}
Finally, the prior $p(\fvm Z)$ is also chosen to be simple, $p(\fvm Z) = \mathcal N (0, \fvm I)$; defining $\widehat{\fvm C} = \ffunc g_\theta (\fvm Z)$ as the reconstructed window, we have the second objective of our entire model:\\[-1cm]

\tikz[remember picture, overlay]{\node[anchor=north, xshift=.46\linewidth, rounded corners=1mm, yshift=-.3cm, draw, minimum height=1.5cm, minimum width=10.5cm, line width=1.5pt, opacity=.8] {};}%
\setlength{\abovedisplayskip}{12pt} %
\setlength{\belowdisplayskip}{12pt} %
\begin{empheq}{align}
\ffunc{L}_\mathrm{ELBO}(\fvm{C}_t) &= \|\fvm{C}_t - \widehat{\fvm{C}}_t\|^2
 + \frac{1}{2L} \sum_{l=1}^{L}\sum_{k=1}^{M}
\left[ \bm\mu_{l}^2 + \bm\sigma_{l}^2 - \log \bm\sigma_{l}^2 - 1 \right]_{k},
\label{eq:elbo_gauss2}
\end{empheq}

\noindent where $[\cdot]_k$ denotes the value at the $k^{th}$ dimension of the vector.
Note that we estimate the couple of parameters $(\bm\mu,\,\bm\sigma)$ for each contextualised time-step in the window.
Thus, we consider a set of parameters $\fset P_t = \{(\bm\mu_l,\,\bm\sigma_l)\}_{l=1}^L$ for each window, which, on average, should follow the same distribution as the prior $p(\fvm Z)$.
We refer to this part of the network as $\textrm{P}_n$ in \Cref{fig:overall}.

\subsubsection{Anomalies from $q_\phi(Z \mid C)$.}
\label{subsec:anomalies}

While the first component of the perturbator, $\textrm{P}_n$, models the distribution of normal windows $p(\fvm{C})$, we introduce a parallel component $\textrm{P}_a$ to infer the anomalous distribution $p(\widetilde{\fvm{C}})$.
Since these distributions are correlated, instead of directly estimating $p(\widetilde{\fvm{C}}\mid\fvm{C})$, we leverage the learned latent estimator $q_\phi(\fvm{Z}\mid\fvm{C})$ and model $p_\psi(\widetilde{\fvm{C}}\mid\fvm{Z})$, parametrised by $\psi$.
Sampling from $q_\phi(\fvm{Z}\mid\fvm{C})$ naturally induces diversity in the generated anomalies.

As no anomalous examples are available, $p_\psi(\widetilde{\fvm{C}}\mid\fvm{Z})$ cannot be learnt generatively like $\ffunc{g}_\theta$.
Instead, inspired by adversarial learning~\cite{gan}, we optimise $\textrm{P}_a$ through the classification objective using a reverse gradient,
$\nabla_{\textrm{P}_a} = -\frac{\partial\ffunc{L}_{CE}}{\partial\fvm{\widetilde{C}}_t}$.
In this case, however, the learnt distribution $p(\widetilde{\fvm{C}}\mid\fvm{Z})$ is approximated only locally around the decision boundary.
The gradient $\nabla_{\textrm{P}_a}$ vanishes for samples already confidently classified as anomalous, but these distant anomalies are trivial and do not contribute meaningful information; hence, we consider this acceptable.

\subsubsection{Architecture.}
The perturbator is built around the Transformer architecture with an innovative configuration.
The encoder $q_\phi$ parametrises a latent distribution of $\fvm C$; the decoders ($\ffunc{g}_\theta$ and $p_\psi$) are conditioned on the resulting sampled latent window $\fvm{Z}^{(i)}$ via cross-attention.
To ensure that information is conveyed through the latent $\fvm Z$, \ie, the reconstruction and generation processes depend on the latent representation, we use positional encodings alone as the decoders inputs.
Full architectural details and implementation specifics are provided in the supplementary material.

\ifSubfilesClassLoaded\clearpage

\end{document}

\section{Experiments}
\label{sec:exp}

In this section, we describe the datasets used in our experiments, outline the baseline methods for comparison.
We then report the experimental results and conclude with an ablation study that evaluates the contribution of each component of our approach.

\subsection{Datasets}

We experimented with four popular multi-variate time-series anomaly detection datasets to validate the effectiveness of {\ours}. The datasets encompass different domains; specifically, they are: 1) Pooled Server Metrics (PSM)~\cite{abdulaal2021practical}, 2) PUMP~\cite{feng2021time}, 3) Secure Water Treatment (SWaT)~\cite{mathur2016swat}, and 4) Controlled Anomalies Time Series (CATSv2)~\cite{fleith2023controlled}.
The statistics of the datasets can be found in supplementary. 
The curated version of the CATSv2 dataset in the TAB benchmark contains 4 subsets, and we train and test on each one. The final results are presented by averaging the metrics for all the subsets.

\subsection{Experimental Setup}

\subsubsection{Metrics.}

We evaluate our method using five metrics: one threshold-based metric, the $\textrm{F}_1$-score, and four rank-based metrics, AUROC, AUPR, VUS-AUROC, and VUS-AUPR.
The Volume Under the Surface (VUS) metrics~\cite{paparrizos2022volume} extend traditional measures into a three-dimensional space by incorporating a temporal buffer size, which accounts for slight early or late detections.
While AUROC and AUPR evaluate performance across thresholds, they are sensitive to temporal misalignments and ignore the range-based nature of anomalies.
VUS calculates the area under ROC or PR curves across all buffer sizes up to the maximum anomaly length, producing a threshold-independent volume that captures a model’s robustness to noise, shifts and timing variations.

For the $\textrm{F}_1$-score, an optimal threshold is needed to convert scores into binary labels.
In TAB, ten percentile values from the classifier outputs on train and test sets are assessed, and the percentile that maximises the score is used as the threshold.
Point Adjustment (PA)-based metrics, commonly used in TSAD studies, are excluded here due to their tendency to inflate performance and obscure true temporal precision~\cite{kim2023contrastive,darban2025carla,kim2022towards,ky2024cats}.
\subsubsection{Baselines.}
We compared {\ours} with 11 baseline methods, which were fairly evaluated under the same protocol in the most recent AD benchmark, TAB~\cite{qiu2025tab}.
We group them  into three different categories based on the main approaches they are using; \textbf{Deep Learning-based} methods, DAGMM \cite{Zong2018DeepAG}, TimesNet~\cite{wutimesnet}, and TranAD~\cite{tuli2022tranad}, large \textbf{Pre-Trained Models}, Timer~\cite{liu2024timer}, TimesFM~\cite{das2024decoder}, Moment~\cite{goswami2024moment}, and Chronos~\cite{ansarichronos}; \textbf{LLM-based} methods, GPT4TS~\cite{zhou2023one}, UniTime~\cite{liu2024unitime}, CALF~\cite{liu2025calf}, and LLMMixer~\cite{kowsher2025llm}.
We directly use the results for the baselines from the TAB~\cite{qiu2025tab} benchmark without any retraining or inference.

\begin{table*}[t]
\centering
\caption{Comparison of TSAD methods over 4 datasets and 5 metrics.
Baseline results are taken from TAB~\cite{qiu2025tab}, and our results are computed using the same benchmark.
\textbf{Best} and \underline{second best} are bold-faced and underlined, respectively.}
\label{tab:main_results}
\resizebox{\textwidth}{!}{%

\begin{tabular}{cl ccc cccc cccc |c}
\toprule

\multirow{3}{*}{Data} &
\multirow{3}{*}{Metrics} & 





 \multicolumn{3}{c}{Deep Learning-based} & \multicolumn{4}{c}{Pre-Trained Models} & \multicolumn{4}{c}{LLM-based}  \\
\cmidrule(lr){3-5} \cmidrule(lr){6-9} \cmidrule(lr){10-14} 

& & 
\multicolumn{1}{c}{DAGMM} & 
\multicolumn{1}{c}{TimesNet} & 
\multicolumn{1}{c}{TranAD}  &

\multicolumn{1}{c}{Timer} & 
\multicolumn{1}{c}{TimesFM}    &
\multicolumn{1}{c}{Moment}  &
\multicolumn{1}{c}{Chronos}    &

\multicolumn{1}{c}{GPT4TS}     & 
\multicolumn{1}{c}{UniTime}   & 
\multicolumn{1}{c}{CALF}       & 
\multicolumn{1}{c}{LLMMixer}  & 

\multicolumn{1}{c}{\textbf{\ours}}  \\

\toprule

\multirow{5}{*}{\rotatebox[origin=c]{90}{PSM}}
&  F$_1$    & 0.007 & 0.088 & \underline{0.403} & 0.143 & 0.082 & 0.117 & 0.131 & 0.089 & 0.097 & 0.102 & 0.186 & \textbf{0.512} \\
&  AUROC      & \underline{0.637} & 0.592 & 0.631 & 0.556 & -     & 0.545 & 0.565 & 0.580 & 0.579 & 0.589 & 0.593 & \textbf{0.697} \\
&  AUPR       & \underline{0.416} & 0.391 & 0.388 & 0.359 & -     & 0.332 & 0.370 & 0.376 & 0.377 & 0.370 & 0.385 & \textbf{0.501} \\
&  VUS-ROC    & \underline{0.608} & 0.593 & 0.566 & 0.541 & -     & 0.540 & 0.547 & 0.576 & 0.579 & 0.587 & 0.579 & \textbf{0.631} \\
&  VUS-PR     & 0.404 & 0.395 & \underline{0.431} & 0.350 & -     & 0.329 & 0.362 & 0.374 & 0.377 & 0.374 & 0.383 & \textbf{0.452} \\

\midrule

\multirow{5}{*}{\rotatebox[origin=c]{90}{PUMP}}
&  F$_1$    & 0.174 & 0.017 & \underline{0.454} & 0.129 & 0.037 & 0.140 & 0.102 & 0.014 & 0.020 & 0.017 & 0.081 & \textbf{0.458} \\
&  AUROC      & 0.447 & 0.485 & \underline{0.800} & 0.529 & 0.409 & 0.485 & 0.469 & 0.414 & 0.549 & 0.566 & 0.398 & \textbf{0.839} \\
&  AUPR       & \underline{0.163} & 0.121 & 0.110 & 0.139 & 0.117 & 0.129 & 0.108 & 0.101 & 0.129 & 0.130 & 0.104 & \textbf{0.254} \\
&  VUS-ROC    & 0.504 & 0.630 & \underline{0.792} & 0.712 & 0.568 & 0.669 & 0.576 & 0.551 & 0.690 & 0.665 & 0.545 & \textbf{0.833} \\
&  VUS-PR     & 0.235 & 0.219 & 0.243 & \underline{0.260} & 0.203 & 0.240 & 0.175 & 0.177 & 0.224 & 0.204 & 0.181 & \textbf{0.290} \\

\midrule

\multirow{5}{*}{\rotatebox[origin=c]{90}{SWAT}}
&  F$_1$    & 0.073 & 0.077 & 0.321 & 0.144 & 0.099 & 0.132 & \underline{0.613} & 0.034 & 0.063 & 0.075 & 0.098 & \textbf{0.695} \\
&  AUROC      & 0.290 & 0.288 & \underline{0.818} & 0.286 & 0.242 & 0.263 & 0.803 & 0.224 & 0.234 &  - & 0.240 & \textbf{0.823} \\
&  AUPR       & 0.082 & 0.107 & \textbf{0.729} & 0.119 & 0.086 & 0.109 & 0.410 & 0.081  & 0.082 & - & 0.086 & \underline{0.542} \\
&  VUS-ROC    & 0.328 & 0.392 & 0.699 & 0.399 & 0.344 & 0.370 & \textbf{0.814} & 0.322 & 0.335 & - & 0.342 & \underline{0.757} \\
&  VUS-PR     & 0.250 & 0.169 & \textbf{0.520} & 0.172 & 0.121 & 0.161 & 0.449 & 0.112 & 0.115 & - & 0.120  & \underline{0.506} \\

\midrule

\multirow{5}{*}{\rotatebox[origin=c]{90}{CATSv2}}
&  F$_1$ & 0.088 & \textbf{0.262} & 0.126 & - & 0.129 & 0.060 & 0.120 & 0.089 & - & 0.138 & 0.081 & \underline{0.172} \\
&  AUROC & 0.619 & \textbf{0.712} & 0.604 & 0.541 & 0.633 & - & 0.624 & 0.601 & 0.334 & 0.616 & 0.617 & \underline{0.665} \\
&  AUPR & 0.090 & \textbf{0.261} & 0.129  & 0.112 &  0.109 & - & 0.103 & 0.094 & 0.039 & \underline{0.133} & 0.085 & 0.106 \\
&  VUS-ROC & 0.667 & \textbf{0.772} & 0.625 & 0.584 & 0.705 & - & 0.705 & 0.676 & 0.373 & 0.689 & 0.689 & \underline{0.709} \\
&  VUS-PR & \textbf{0.314} & \underline{0.265} & 0.130 & 0.132 & 0.125 & - & 0.112 & 0.107 & 0.058 & 0.145 & 0.099 & 0.110 \\


\bottomrule

\end{tabular}

 }
\end{table*}

\subsection{Results}

We present the results of {\ours} and the selected baselines in Table~\ref{tab:main_results}.
Comparing {\ours} to previous LLM-based methods, our approach significantly improves upon all metrics, except for the CATSv2 dataset.
For instance, on the PSM and SWaT datasets, the $\textrm{F}_1$-score is improved by more than $0.3$ points, which demonstrates the effectiveness of our method.
Compared to all baselines, on the PSM and PUMP datasets, {\ours} marks a new state-of-the-art (SoTA) across all metrics.
On the SWAT dataset, our method achieves SoTA results on $\textrm{F}_1$ and AUROC, and ranks second on the other three metrics.
Finally, on the challenging CATSv2 dataset, which contains very few anomalous regions, our method achieves the second best scores on a variety of metrics, following TimesNet~\cite{wutimesnet}.

\subsection{Ablation Study}

\begin{table*}[t]
\centering

\begin{minipage}{0.49\textwidth}
    \centering
    \captionof{table}{\ours{} with and without an LLM and different fine-tuning schemes.}
    \label{tab:abl_llm}
    \resizebox{\linewidth}{!}{
\begin{tabular}{cc ccc ccc}
\toprule
\multirow{3}{*}{LLM} & \multirow{3}{*}{\shortstack{Fine \\ Tuning}} & \multicolumn{3}{c}{PSM} & \multicolumn{3}{c}{PUMP} \\ 
\cmidrule(lr){3-5} \cmidrule(lr){6-8} 
& & F$_1$ & AUROC & AUPR & F$_1$ & AUROC & AUPR \\
\toprule
\XSolidBrush & \XSolidBrush             & 0.479 & 0.669 & 0.447     & 0.021  & 0.460 & 0.087 \\
GPT-2        & \XSolidBrush             & 0.354 & 0.496 & 0.389     & 0.370  & 0.802 & 0.199 \\
GPT-2        & \Checkmark               & 0.361 & 0.499 & 0.417     & 0.265  & 0.707 & 0.146 \\
\midrule
GPT-2        & \Checkmark (LoRA)        & 0.512 & 0.697 & 0.501     & 0.458  & 0.839 & 0.254\\
\bottomrule
\end{tabular}

}
\end{minipage}
\hfill
\begin{minipage}{0.49\textwidth}
    \centering
    \vspace{-1.55em}
    \captionof{table}{\ours{} and its Transformer-based classifier \textit{vs}. an MLP-based one.}
    \label{tab:abl_classifier}
    \resizebox{\linewidth}{!}{
\begin{tabular}{c ccc ccc}
\toprule
\multirow{3}{*}{Classifier} & \multicolumn{3}{c}{PSM} & \multicolumn{3}{c}{PUMP} \\ 
\cmidrule(lr){2-4} \cmidrule(lr){5-7} 
 & F$_1$ & AUROC & AUPR & F$_1$ & AUROC & AUPR \\
\toprule
Transformer  & 0.512 & 0.697 & 0.501 & 0.458  & 0.839 &  0.254    \\
\midrule
MLP-based    & 0.497 & 0.724 & 0.471 & 0.016  & 0.307 &  0.069    \\
\bottomrule
\end{tabular}

}
\end{minipage}

\end{table*}

We conduct ablation experiments on the PSM and PUMP datasets to evaluate different components of {\ours}.
For the LLM and fine-tuning, we compare four feature-extractor setups: a linear layer only, a frozen LLM, standard LLM fine-tuning~\cite{zhou2023one}, and LoRA fine-tuning (our approach).
Using only the linear layer is already competitive, highlighting the effectiveness of the classifier and perturbator. Adding a frozen LLM improves AUROC on PUMP, demonstrating the benefit of contextualisation.
Standard fine-tuning slightly helps PSM but harms PUMP, while LoRA achieves the best results on both datasets, showing its superiority for leveraging LLMs in TSAD (Table~\ref{tab:abl_llm}).

To evaluate the impact of the proposed classifier architecture, we replace it with an MLP while maintaining a constant number of trainable parameters.
The MLP-based classifier yields inferior performance, especially on the PUMP dataset, highlighting the importance of modelling the sequential nature of the time series (Table~\ref{tab:abl_classifier}).

\ifSubfilesClassLoaded\maketitle

\section{Discussion \&  Future Work}
\label{sec:discussion}

We analyse the perturbator components $\textrm{P}_n$ and $\textrm{P}_a$, and discuss the limitations and future directions (figures in supplementary).
To ensure VAE non-collapse, we monitored $\bm\mu_l$ and $\bm\sigma_l$ and observed convergence near $0$ and $1$ with deviations $\sim10^{-5}$–$10^{-6}$, while the reconstruction loss decreased alongside $\ffunc D_{KL}$, indicating that sufficient information is transmitted.
Cosine distances between generated pseudo-anomalies $\widetilde{\fvm C}_t$ and normal windows $\fvm C_t$ decrease during training, indicating that pseudo-anomalies become more complex while still being distinguishable by the classifier.

We observed that the generated pseudo-anomalies are sometimes insufficiently diverse; however, they remain closer to real anomalies than to normal data, which aids the classifier in learning meaningful boundaries.
Future work may enhance diversity through complementary latent sampling, relaxed prior constraints (\eg, MMD), or distributional constraints on $\textrm{P}_a$.
Even though \ours{} converges properly, we also monitored large magnitude differences in its loss components, indicating that the objectives are not being optimised equally and suggesting potential benefits from balancing strategies.
Finally, the linear layer used to map time-series to token embeddings may be restrictive, and non-linear alternatives such as MLPs or Kolmogorov–Arnold Networks~\cite{Liu2024KANKN} could manage the complex alignment better.

We note that evaluating on more datasets would better assess \ours{} performance, given the limitations and heterogeneity of multivariate TSAD data~\cite{Wu2020CurrentTS}.

\ifSubfilesClassLoaded\clearpage

\end{document}

\section{Conclusion}
\label{sec:conc}

In this work, we tackle unsupervised TSAD and overcome the limitations of handcrafted, dataset-specific pseudo-anomalies by introducing ASTER, the first framework to generate pseudo-anomalies autonomously in a learnt latent space.
Leveraging pre-trained LLM representations, the expressive capacity of Transformers, and the modelling power of VAEs, ASTER produces challenging pseudo-anomalies to adversarially train a robust Transformer-based classifier.
Extensive experiments on the TAB benchmark demonstrate that ASTER achieves state-of-the-art performance across datasets, outperforming existing LLM-based methods.
Future research could focus on further enhancing the perturbator by enforcing anomaly diversity through advanced latent sampling, constraints, or more complex decoder inputs (\eg, learnable tokens), exploring the interpretability of the latent space and the generated pseudo-anomalies, and investigating various pre-trained LLMs and their differing representational performance with enhanced time-series-to-tokens translation modules.

\subsubsection{Acknowledgements.}
This research was funded by the Luxembourg National Research Fund (FNR), grant reference \texttt{DEFENCE22/17813724/AUREA}.
%

%
%
%
\bibliography{bib2,general}
\clearpage
\appendix

\title{ASTER: Latent Pseudo-Anomaly Generation for Unsupervised Time-Series Anomaly Detection}
\maketitlesupplementary

\section{Complementary Figure: Positioning}

\begin{figure}
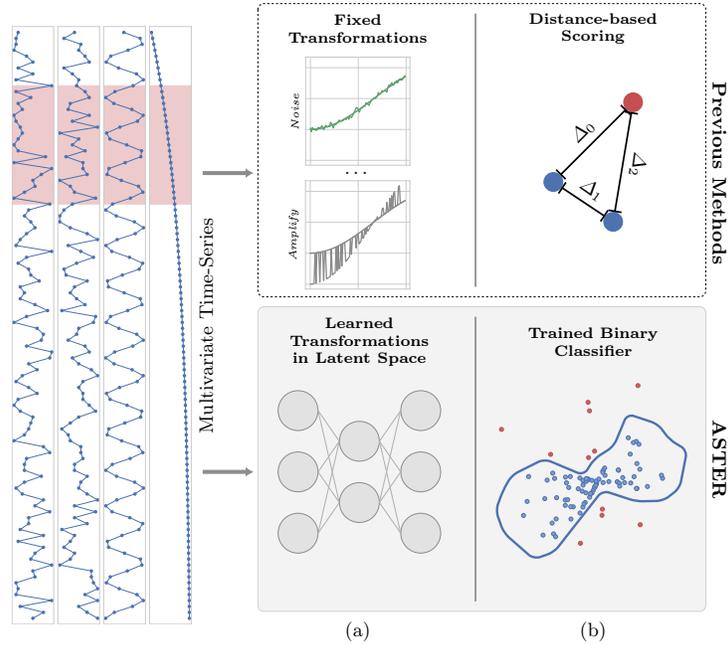

    \centering
    \vspace{-1cm}
    \resizebox{.8\textwidth}{!}{\subfile{../figures/tikz/teaser}}
    \caption{Main differences between previous methods and \ours{}; (a) the pseudo-anomaly generation, (b) the scoring technique. Previous methods curate a list of augmentations to simulate the possible anomalies that exist in the time-series datasets. However, our method is free from these assumptions and learn to generate the pseudo-anomalies automatically in the latent space. Also, the training objective of these methods is to maximise the distance between pseudo-anomalies and normal data. This approach results in specialised distance metrics (\eg, cosine, $L_2$) for each dataset based on the characteristics. {\ours} relaxes this by training a binary classifier. \textcolor{color3}{Red} and \textcolor{color0}{blue} dots denote anomalies and normal data, respectively.}
    \label{fig:teaser}
\end{figure}

\section{Datasets Statistics \& Processing}

The statistics of the datasets studied in this work, mainly, their size and percentage of anomalies, are presented in \Cref{tab:datasets}.

We use standard scaling, \ie, centering the features by removing the mean and scaling them to unit variance as a pre-processing step on all datasets.
The mean and variance are computed solely on the training set.

\begin{table}[H]
    \caption{Details of the datasets used in experiments.}
    \centering

\begin{tabular}{lcccc}
\toprule
\textbf{Dataset} & \textbf{Train} & \textbf{Test} & \textbf{Features} & \textbf{Anomalies (\%)} \\
\midrule
PSM \cite{abdulaal2021practical}  & 132,481 & 87,841 & 25 & 27.76 \\
PUMP~\cite{feng2021time} & 76,901 & 143,401  & 44 & 10.05 \\
SWaT~\cite{mathur2016swat} & 495,000 & 449,919  & 51 & 12.14 \\
CATSv2~\cite{fleith2023controlled} & 113,887 & 576,113  & 17 & 3.21 \\

\bottomrule
\end{tabular}

    \label{tab:datasets}
\end{table}

\section{Results Origin}
The results of the main paper have been taken directly from the TAB~\cite{qiu2025tab} GitHub repository (\url{https://github.com/decisionintelligence/TAB}).
Since the writing of the paper, the results tables have been removed from the main branch.
However, they can be found back on commit ``\texttt{d666a9796412c17619e599308cdff 9edfa619e6a}'', in ``\texttt{docs/results/TAB multivariate dataset result.pdf}''.

The latest results for most of the compared methods can be found on their new benchmark page (\url{https://decisionintelligence.github.io/OpenTS}).

\section{Algorithms}

We present the main steps of \ours's inference algorithm in \Cref{alg:eval}.
The training algorithm is depicted in \Cref{alg:train}; the steps during back-propagation are referred to as \textit{BP}.

Specifically, to train $\textrm{P}_a$ we use the reverse gradient operation (\Cref{alg:train} L-15).
Before back-propagating the gradients coming from the classifier into the pseudo-anomaly generator $\textrm{P}_a$ to update its weights, we scale them by $-1$.
This has the effect of inverting the classification objective to train $\textrm{P}_a$, thus training $\textrm{P}_a$ to make the classification task between normal and pseudo-anomalous samples more challenging.

We then stop the back-propagation of the gradients coming from $\textrm{P}_a$ by scaling them by $0$ (\Cref{alg:train} L-13).
This strategy allows the updates $q_{\phi}$'s parameters (encoder) not to be affected by the pseudo-anomaly generation.
We apply the same stop gradient strategy for the gradients coming from $q_{\phi}$ (\Cref{alg:train} L-4).
This way, the learnable weights of $\Phi$ (adapter $\Phi_0$ and fine-tuning parameters of the LLM $\Phi_1$) are updated solely from the normal samples classification results.

\begin{algorithm}[t]
\caption{Inference Pipeline of {\ours}}
\label{alg:eval}
\begin{algorithmic}[1]
\Require\\
\begin{itemize}
    \item Unseen input window $\fvm W^*_t$
    \item Feature extractor $\Phi$
    \item Classifier $\Psi$
\end{itemize}
\algsection{Feature Extraction}
\State $\fvm C^*_t \gets \Phi(\fvm W^*_t)$
\algsection{Scoring (Classification)}
\State $s^*_t = \Psi(\fvm C^*_t)$

\end{algorithmic}
\end{algorithm}

\begin{algorithm}[H]
\caption{Training Procedure of {\ours}}
\label{alg:train}
\begin{algorithmic}[1]
\Require\\
\begin{itemize}
    \item Input window and its size $\fvm W_t$, $L$
    \item Auto-encoder $\textrm{P}_n$ ($q_\phi$, $\ffunc g_\theta$)
    \item Pseudo-anomaly generator $\textrm{P}_a$ ($p_\psi$)
    \item Feature extractor $\Phi$
    \item Classifier $\Psi$
    \item Positional encodings $\fvm{PE}$
\end{itemize}
\algsection{Feature Extraction}
\State{$\fvm C_t \gets \Phi(\fvm W_t)$}
\State $\fvm C_t' \gets \fvm C_t $
\Comment{\textit{BP--Stop Gradient}}
\algsection{Encoding \& Sampling}
\State $\fset P_t \gets q_\phi(\fvm C'_t +\fvm{PE})$
\For{$l=1$ to $L$}
    \State $\bm\epsilon_l^{(i)} \sim \mathcal N(0, \fvm I)$
    \State $\fvm Z^{(i)}_{[l]} \gets [\bm \mu_l + \bm \sigma_l \odot \bm\epsilon_l^{(i)}]$
    \Comment{$(\bm\mu_l,\,\bm\sigma_l) \in \mathcal P_t$}
\EndFor
\algsection{Reconstruction \& Generation}
\State $\widehat{\fvm C}_t \gets \ffunc g_\theta (\fvm Z^{(i)};\;\fvm{PE} )$
\Comment{As: (Cross-Attention Inputs; Inputs)}
\State $\fvm{Z}'^{(i)} \gets \fvm Z^{(i)}$
\Comment{\textit{BP--Stop Gradient}}
\State{$\widetilde{\fvm C}_t \gets p_\psi (\fvm Z'^{(i)};\;\fvm{PE} )$}
\State $\widetilde{\fvm C}'_t \gets \widetilde{\fvm C}_t$
\Comment{\textit{BP--Reverse Gradient}}
\algsection{Classification}
\State $\Bar{s}_t \gets \textrm{Sigmoid}(\Psi(\fvm C_t))$
\State $\Bar{\tilde{s}}_t \gets \textrm{Sigmoid}(\Psi(\widetilde{\fvm C'}_t))$
\algsection{Losses Computation}
\State{$\ffunc L_\textrm{CE} \gets -\left[\mathrm{log}(1 - \Bar{s}_t) + \mathrm{log}(\Bar{\tilde{s}}_t)\right]$}
\State{$\ffunc L_\textrm{ELBO} \gets MSE(\fvm C_t, \widehat{\fvm C}_t) + \ffunc D_\textrm{KL}(\fset P_t)$}
\end{algorithmic}
\end{algorithm}

\section{Architectural Details}
\subsection{Perturbator}
We use the Transformer architecture~\cite{vaswani2017attention} at the core of $\textrm{P}$.
While $q_\phi$ is based on the encoder part, $\ffunc g_\theta$ and $p_\psi$ are based on the decoder part, \ie, with cross-attention heads.

The encoder, $q_\phi$ is divided into three parts: $q_{\phi,0}$, the actual Transformer structure; $q_{\phi,\mu}$, a linear layer encoding each time-step $l$ into its distribution's mean $\bm\mu_l$; $q_{\phi,\sigma}$, a linear layer encoding each time-step $l$ into its distribution's standard deviation $\sigma_l$.

During training, a sample $\fvm Z^{(i)}$ is drawn from the distribution $q_\phi(\fvm Z\mid\fvm C)$ for each window.
The reparametrisation trick and the auxiliary variable $\bm\epsilon$ are used for each time-step:
\begin{equation}
\fvm Z^{(i)} = \left[\bm\mu_l + \bm\sigma_l\odot\bm\epsilon_l^{(i)}\right]_{l=1}^{L},
\end{equation}
\noindent and, inspired by previous works in music~\cite{Jiang2020TransformerVA} and face~\cite{Zou20233DFE} generation, we used the sampled window as cross-attention keys and values for the decoders $\ffunc g_\theta$ and $p_\psi$.

On top of the cross-attention values, the decoders must be given inputs to determine the query tokens.
Since the objective for $\ffunc g_\theta$ is raw reconstruction, instead of next-token prediction or forecasting, we cannot give shifted inputs or partial inputs to the decoders, at the risk of shifting the meaning of $\fvm Z$.
Neither can we give the $\fvm C_t$, since, at least for $\fvm g_\theta$, this would give a simple solution during optimisation: passing the input directly as output.

Instead, we propose to add positional encoding to $\fvm C_t$ before processing it with $q_\phi$ and, only give those same positions as input to $\ffunc g_\theta$ and $p_\psi$.
This, in fact, allows one to maximise the amount of information flowing through $\fvm Z$.
We use sinusoidal positional encoding~\cite{vaswani2017attention} as formulated below, with $k$ as the feature dimension index:\\[-.2cm]
\begin{equation}
\mathbf{PE}_{t,2k}   = \sin\!\Big(t / 10000^{\frac{2k}{M}}\Big),\qquad
\mathbf{PE}_{t,2k+1} = \cos\!\Big(t / 10000^{\frac{2k}{M}}\Big)
\end{equation}

\subsection{Implementation}
We implemented our method using PyTorch~\cite{pytorch}, and all experiments were conducted on Nvidia A100 GPUs.
We train our model using the SGD optimiser with a fixed learning rate of $0.001$ (without any scheduling), and a batch size of 64.
All models are trained for $100$ epochs, and the last epoch is used for inference.
We select $4$ as the window size.
The depth of the classifier and the perturbator components is set to $2$.
Similar to previous methods, we use the small variant of GPT-2~\cite{radford2019language} as the base LLM.
However, unlike previous methods, we finetune the attention and projection layers using LoRA~\cite{lora}. 
The model has a total of $180$M parameters, with around $55$M trainable. For a fair comparison and reproducible results, we use the default seed in the TAB benchmark for all the experiments.

\subsection{Time \& Memory Complexity}

We compare the complexity of \ours{} to TimesNet~\cite{wutimesnet} and GPT4TS~\cite{zhou2023one} in \Cref{tab:complex}.
Results were obtained on a NVIDIA RTX A4000 graphics cards, batch size of 128.

Comparing the methods complexity \wrt{} their best window size configuration, \ours{} is the fastest but requires more memory than TimesNet.
However, \ours{} scales poorly with an increasing window size due to the number of tokens increasing linearly (one token per time-step).
A solution to limit this behaviour could be to apply patching~\cite{Nie2022ATS} or to change the architecture of the learnable projection layer $\Phi_0$ to include time-wise dimension reduction.
Moreover, most of the complexity comes from $\Phi_1$, the chosen LLM; thus, an architecture optimised for handling long token sequences would also be more competitive.

\begin{table}[h]
    \centering
    \caption{Comparison of time (one epoch) and memory (maximum usage) costs of \ours{} and two other baselines, on SWAT dataset. Configurations (\wrt{} Window Size) achieving the best results are highlighted in grey.}
    \resizebox{.8\linewidth}{!}{
\begin{tabular}{cccccc}
\toprule
\multirow{2}{*}{\textbf{Method}} & \multirow{2}{*}{\textbf{Window Size}} &\multicolumn{3}{c}{\textbf{Time (s)}} & \multirow{2}{*}{\textbf{Memory (MB)}} \\
\cmidrule(lr){3-5}
&& \textbf{Train} & \textbf{Validation} & \textbf{Test} &\\
\midrule
\multirow{5}{*}{\ours{}} & \cellcolor{gray!20}4 & \cellcolor{gray!20}209 & \cellcolor{gray!20}15 & \cellcolor{gray!20}51 & \cellcolor{gray!20}2572\\
& 8 & 336& 25& 86& 3523\\
& 16 & 588& 45& 152& 5364 \\
& 32 & 1083& 83& 285& 8845 \\
& 64 & 2074& 160& 545& 16145\\
TimesNet & \cellcolor{gray!20}100 & \cellcolor{gray!20}274 & \cellcolor{gray!20}22 & \cellcolor{gray!20}98 & \cellcolor{gray!20}951 \\
GPT4TS & \cellcolor{gray!20}100 & \cellcolor{gray!20}997& \cellcolor{gray!20}80& \cellcolor{gray!20}369& \cellcolor{gray!20}9370\\
\bottomrule
\end{tabular}

}
    \label{tab:complex}
\end{table}

\section{Complementary Ablation} 
There is no consensus on the optimal window size in time-series anomaly detection, and most existing methods determine the window length empirically, often relying on heuristic tuning or dataset-specific experimentation.
We conduct experiments to determine the best window size, and the results are presented in Table~\ref{tab:abl_windowsize}.  

\begin{table}[H]
    \centering
    
\begin{tabular}{c ccc ccc}
\toprule
\multirow{3}{*}{\shortstack{Window \\ Size}} & \multicolumn{3}{c}{PSM} & \multicolumn{3}{c}{PUMP} \\ 
\cmidrule(lr){2-4} \cmidrule(lr){5-7} 
 & F$_1$ & AUROC & AUPR & F$_1$ & AUROC & AUPR \\
\toprule
4  & 0.512 & 0.697 & 0.501 & 0.458  & 0.839 &  0.254    \\
\midrule
8  & 0.456 & 0.644 & 0.387 & 0.399  & 0.774 &  0.185    \\
16 & 0.450 & 0.639 & 0.387 & 0.429  & 0.777 &  0.189     \\
32 & 0.461 & 0.656 & 0.419 & 0.433  & 0.785 &  0.191    \\
64 & 0.475 & 0.681 & 0.487 & 0.441  & 0.828 &  0.224    \\
\bottomrule
\end{tabular}

    \caption{Ablation experiments to determine optimum window size. We observe window size of 4 gives the best performance. Increasing the window size further from $8$ to $64$ gives a gradual improvement with the cost of quadratic increase on the runtime due to the attention-based modules.}
    \label{tab:abl_windowsize}
\end{table}

\section{Model Analysis: Visualisation}
\begin{figure}[H]
    \centering

    \begin{minipage}[t]{0.58\linewidth}
        \centering
        \includegraphics[width=\linewidth]{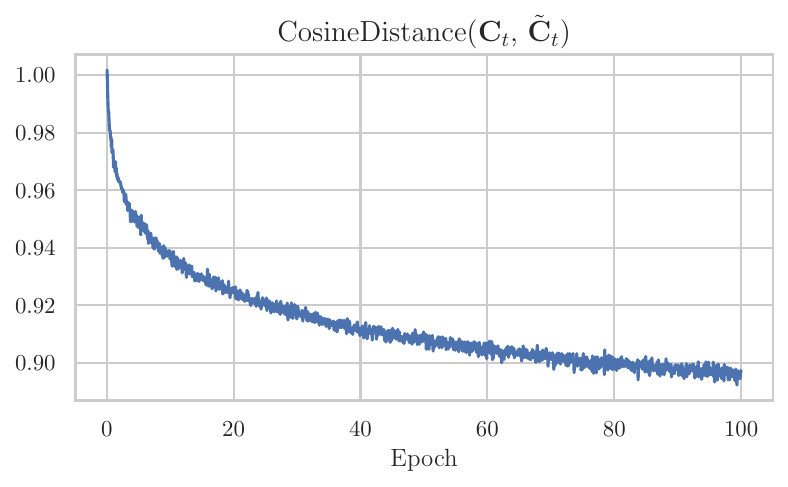}
        \caption{The cosine distance between the real and pseudo-anomaly data during training. The decreasing distance indicates that the perturbator generates increasingly challenging pseudo-anomalies throughout training.}
        \label{fig:pa-dist}
    \end{minipage}
    \hfill
    \begin{minipage}[t]{0.38\linewidth}
        \centering
        \includegraphics[width=\linewidth]{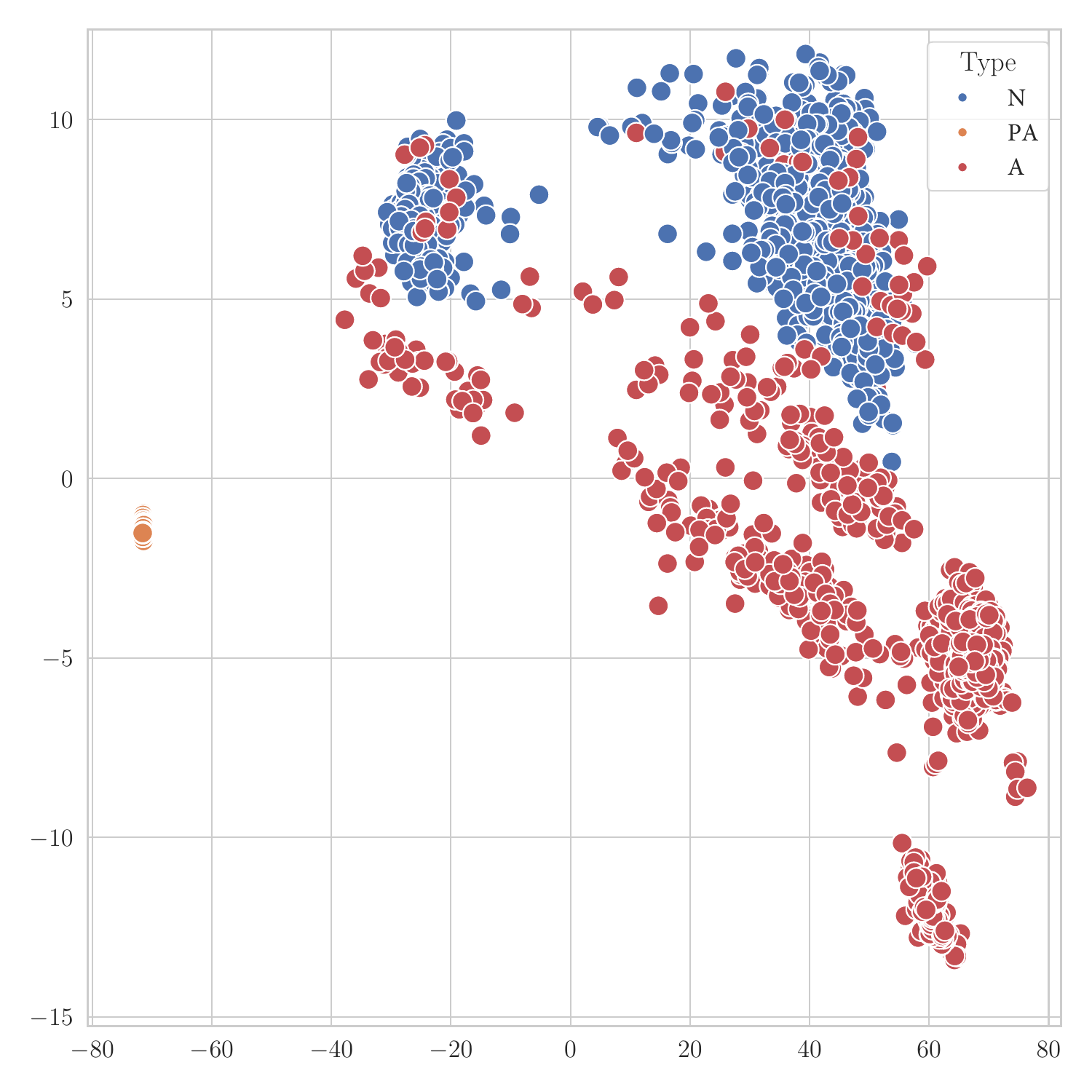}
        \caption{PCA visualisation of normal (N), real anomaly (A), and generated pseudo-anomaly (PA) data. Features are extracted before the classifier, resulting in an unclear boundary between normal and anomalous data at this stage.}
        \label{fig:pseudo_tsne}
    \end{minipage}

\end{figure}

We computed the average cosine distance (\Cref{eq:cosinedist}) between the generated pseudo-anomaly $\widetilde{\fvm C}_t$ and the normal window $\fvm C_t$ throughout the training.
The resulting plot (\Cref{fig:pa-dist}) suggests that pseudo-anomalies become increasingly entangled with normal windows, making them more difficult to discriminate.
Still, our observations indicate that the classifier continues to distinguish these windows throughout training, highlighting the effectiveness of the learnt generation process.

\begin{equation}
    \label{eq:cosinedist}
    \text{CosineDistance}(\fvm C_t, \widetilde{\fvm C}_t)
    = 1 - \frac{1}{L} \sum_{l=1}^{L}
    \left[
        \frac{
            \fvm C_{t,l} \cdot \widetilde{\fvm C}_{t,l}
        }{
            \| \fvm C_{t,l} \|_2 \, \| \widetilde{\fvm C}_{t,l} \|_2
        }
    \right].
\end{equation}

In Figure~\ref{fig:pseudo_tsne}, we present the PCA plots of the distributions of normal (N), real anomaly (A), and pseudo-anomaly (PA) data on the test set of the PSM dataset.
We observe that the diversity of pseudo-anomalies is limited--at least in the projected dimensions where normal and anomalous data seem to be very sparse.
This points to a promising direction for further enhancing Requirement 3 and achieving improved results.
However, the pseudo-anomalies seem ``closer'' to the real anomalies than to the normal data, and as shown in Figure~\ref{fig:pa-dist}, the distance between the normal data and the pseudo-anomalies is decreasing.
These results indicate that even a limited diversity of pseudo-anomalies assists the model in learning a meaningful classification boundary.

%
\end{document}